\documentclass[letterpaper, 10 pt, conference]{ieeeconf}  

\IEEEoverridecommandlockouts                              

\usepackage{amsmath}

\usepackage{graphicx}
\usepackage{wrapfig} 
\usepackage[noadjust	]{cite}
\usepackage{amssymb}

\usepackage{mathtools} 
\usepackage{bm}
\usepackage{isomath} 
\usepackage{csquotes} 
\usepackage{acro} 
\usepackage{booktabs} 
\usepackage{algorithm}
\usepackage{algorithmic}
\usepackage{caption}
\usepackage{subcaption}
\usepackage{booktabs}
\usepackage[table]{xcolor}
\usepackage{multirow}
\usepackage{hhline}
\usepackage{pifont}
\newcommand{\cmark}{\ding{51}}

\newcommand{\vc}[1]{\bm{#1}} 
\newcommand{\mat}[1]{\bm{#1}} 

\newcommand{\R}{\mathbb{R}}  

\DeclarePairedDelimiter{\norm}{\lVert}{\rVert} 

\ifdefined\briefref

    \newcommand{\eqlower}{eq.}

\else

    \newcommand{\eqlower}{equation}

\fi

\let\oldeqref\eqref
\renewcommand{\eqref}[1]{\eqlower~\oldeqref{#1}}  

\newcommand{\img}[1]{\mat{I}_{#1}}
\newcommand{\imga}{\img{A}}
\newcommand{\imgaaug}{\img{\hat{A}}}
\newcommand{\imgb}{\img{B}}

\newcommand{\dimg}[1]{\mat{D}_{#1}}
\newcommand{\dimga}{\dimg{A}}
\newcommand{\dimgb}{\dimg{B}}

\newcommand{\kp}[1]{\vc{k}_{#1}}
\newcommand{\kpi}[2]{\vc{k}_{#1}^{#2}}
\newcommand{\kpf}[1]{\vc{k}_{#1} = (x_{#1}, y_{#1})}

\newcommand{\desc}[1]{\vc{d}_{#1}}

\newcommand{\CCL}{CCL} 

\usepackage{color}
\newcommand{\todo}[1]{}
\renewcommand{\todo}[1]{{\color{red} TODO: {#1}}}

\title{\LARGE \bf
Cycle-Correspondence Loss: Learning Dense View-Invariant Visual Features from Unlabeled and Unordered RGB Images
}

\author{David B. Adrian$^{1,2}$, Andras Gabor Kupcsik$^{1}$, Markus Spies$^{1}$, Heiko Neumann$^{2}$
	\thanks{$^{1}$ Bosch Center for Artificial Intelligence, Renningen, Germany, 
		{\tt\small firstname(s).lastname@de.bosch.com}}%
	\thanks{$^{2}$  Institute of Neural Information Processing, Ulm University, Ulm, Germany,
		{\tt\small firstname.lastname@uni-ulm.de}}%
}

\begin{document}

\maketitle
\thispagestyle{empty}
\pagestyle{empty}

\begin{abstract}

Robot manipulation relying on learned object-centric descriptors became popular in recent years.
Visual descriptors can easily describe manipulation task objectives, they can be learned efficiently using self-supervision, and they can encode actuated and even non-rigid objects.
However, learning robust, view-invariant keypoints in a self-supervised approach requires a meticulous data collection approach involving precise calibration and expert supervision.
In this paper we introduce \emph{Cycle-Correspondence Loss} (\CCL) for view-invariant dense descriptor learning, which adopts the concept of cycle-consistency, enabling a simple data collection pipeline and training on unpaired RGB camera views.
The key idea is to autonomously detect valid pixel correspondences by attempting to use a prediction over a new image to predict the original pixel in the original image, while scaling error terms based on the estimated confidence.
Our evaluation shows that we outperform other self-supervised RGB-only methods, and approach performance of supervised methods, both with respect to keypoint tracking as well as for a robot grasping downstream task.

\end{abstract}


\section{Introduction}

Dense visual descriptors have proven to be a flexible, easy to learn, and easy to use object representation for robot manipulation in recent years.
They show potential for class-level object generalization \cite{Florence2018}, they can describe non-rigid objects \cite{Sundaresan2020}, and they can be seamlessly applied for state-representation for control \cite{Manuelli2019, Florence2020, Manuelli2020}.
A dense descriptor network maps an RGB image of size $3\times H\times W$ to a descriptor space image of size $D\times H \times W$, where $D$ is the user-defined descriptor dimension.

Training a dense descriptor network, such as a Dense Object Net (DON) \cite{Florence2018}, relies on multiple views of the same object(s) and dense pixel correspondences computed from 3D geometry \cite{Florence2018, Yang2021}.
Alternatively, RGB image augmentations can generate alternative views of the same image, while keeping track of pixel correspondences \cite{Thewlis2017, Novotny2018, graf2023learning}.
Training is commonly achieved, e.g., via contrastive \cite{Hadsell2006, Chen2020} or probabilistic~\cite{Florence2020} losses.

Utilizing pixel correspondences computed by 3D geometry naturally encodes physically distinct views of the same object(s), thus encouraging truly view-invariant descriptors.
However, this requires a registered RGB-D dataset \cite{Florence2018} or trained NeRF \cite{Yen-chen22}, which is often laborious due to camera calibration, hardware setup, and data logging.
This is exactly the problem the RGB image augmentation approaches \cite{Thewlis2017, Detone2018, graf2023learning} aim to solve: they only require an unordered set of RGB images depicting the object(s), which can be recorded even with a smartphone.
However, the learned descriptors cannot handle excessive camera view changes \cite{graf2023learning}, and thus, they are not always view-invariant, which limits their applicability.
In this work our aim is to combine the best of both worlds.
Firstly, we wish to keep the simple data collection approach, that is, relying only on a set of unordered RGB images showing the objects.
Secondly, we aim to improve the view-invariance of the descriptors, making them more robust to camera view changes or extreme object positions.

\begin{figure}[t!]
	\centering
	\includegraphics[width=\linewidth]{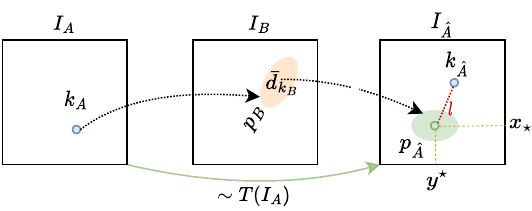}
	\caption{
		Overview of the cycle-correspondence loss.
		$\imga$ and $\imgaaug$ denote versions of the same image, both related through a random image transformation $\thicksim T$.
		$\imgb$ is a randomly sampled image that exhibits partial content overlap with $\imga$.
		We establish a correspondence cycle by randomly sampling location $\kp{A}$ on $\imga$, computing a matching distribution $p_B$ over $\imgb$ which we utilize to predict $\kp{\hat{A}}$ on $\imgaaug$.
		As location $\kp{\hat{A}}$ is known through the augmentation, we can optimize the prediction error $l$ to improve the model.
		We utilize the predicted distributions to scale individual error terms $l$ by the associated uncertainty, effectively dealing with sampled $\kp{A}$ that have no valid correspondence in $\imgb$.
		}
	\label{fig:cycle_overview}
	\vspace{-20pt}
\end{figure}

To this end, we introduce the Cycle-Correspondence Loss (\CCL), a self-supervised loss for dense visual feature models using only unlabled, random pairs of RGB images.
The core idea, based on \textit{cycle-consistency}, is that for an image pair $(I_A, I_B)$, given unique descriptors in image $I_A$ and $I_B$, any correctly predicted keypoint location in image $I_B$ can in turn be used to predict the original point in image $I_A$, completing a \textit{cycle} of correspondence predictions, see Fig. \ref{fig:cycle_overview} for a visual overview.
The model is able to learn by itself to detect valid correspondences, without relying on ground-truth correspondence annotations, by estimating uncertainties and scaling contribution of error terms accordingly.
The only assumption is that the sampled training image pairs at least partially depict the same content with unique object instances.
This still allows for random object arrangements, varying backgrounds, and scene conditions.
Our loss is generally applicable, and can thus also be used with existing annotations, sim-to-real data generation, and other methods.

\section{Related Work}
Keypoint detection from RGB images in robot learning and control has been extensively researched in recent years.
\textit{Sparse keypoint techniques} provide a discrete set of task-relevant keypoint locations in image plane or in camera coordinates.
Early methods exploited autoencoders to reconstruct images with the bottleneck as keypoints \cite{Jakab2018} or keypoint distributions \cite{Zhang2018}, and learned meaningful keypoints for solving ATARI games \cite{Kulkarni2019}.
Following similar ideas, \cite{Suwajanakorn2018} proposes to learn object-category level keypoints from 3D models in a fully self-supervised way.
More relevant to robot manipulation, the KeyPose method proposes to learn sparse keypoints for transparent objects using stereo RGB cameras \cite{Liu2020}.
Manuelli et al. showed that keypoint representation can be efficiently used to solve robot manipulation tasks \cite{Manuelli2019}.
Some of the most promising results with sparse keypoints for robot manipulation using human annotation and self-supervision was shown by Vecerik et al. in \cite{Vecerik2020, Vecerik2022}.

\textit{Dense keypoint methods} predict a single descriptor vector for every pixel of the RGB image.
Florence et al.~\cite{Florence2018} proposed Dense Object Nets (DON) for fully autonomous object-centric dense descriptor learning.
This work inspired a variety of follow up research, such as, applications for behavior cloning \cite{Florence2020}, learning model predictive controllers \cite{Manuelli2020} and even rope manipulation \cite{Sundaresan2020}.
Other works focused, e.g., on better generalization for multiple object classes \cite{Yang2021} or class aware descriptors \cite{Hadjivelichkov2021}.
It has also been shown how to improve the original work \cite{Florence2018} with alternative losses and training regimes \cite{florence2020dense,Adrian2022} and how to avoid costly preprocessing \cite{Adrian2022}.
Recently, Yen-Chen et al.~\cite{Yen-chen22} applied NeRFs to learn DON from registered RGB scenes.

There has been another line of work focusing on learning dense descriptors from RGB images only, without the costly data collection and preprocessing.
In the computer vision community image augmentations have been proposed to generate alternative views of the same image and use self-supervision for learning \cite{Thewlis2017, Novotny2018}.
\cite{graf2023learning} applied similar techniques to the robotics domain and showed that view-invariance of such descriptors are limited.
SuperPoint is a pretrained method that uses a keypoint location heatmap and a dense descriptors head to provide robust keypoint locations \cite{Detone2018}.
Deekshith et al. showed that optical flow from video can also be used to learn dense descriptors \cite{Deekshith2020}.
It is also possible to implicitly train a dense descriptor model through autonomous grasp interactions \cite{JangDVL18}, however, this requires a large amount of grasp interactions to do so.
Another recent, but promising line of research investigates the usage of large pre-trained vision  transformer models~\cite{caron2021emerging,oquab2023dinov2} as provider of off-the-shelf features \cite{amir2021deep}.
For example, Hadjivelichkov et al.~\cite{Hadjivelichkov2021} already demonstrated their usability to obtain one-shot affordance regions for robotic manipulation.

Our work builds on the idea of cycle-consistency, a well-established concept that is used, e.g, in CycleGAN \cite{CycleGanJunYan} for image-to-image translation, for temporal correspondence learning in \cite{temporal_cycle_correspondence}, or correspondence learning via 3D CAD models in \cite{3dcorrespondences_cyle}.
WarpC \cite{TruongWarpC} and PWarpC \cite{TruongPWarpC} utilize cycle-consistency to predict dense flows across two unpaired images and an augmented version that induces a known warp.
Due to the close relation to our model, we explicitly discuss differences to these two models in more detail in Sec. \ref{sec:relation_warpc}.


\section{Method}
\label{sec:method}
In the following, we first outline our notation and preliminary concepts, followed by introducing \CCL{}, see Figure \ref{fig:cycle_overview}, and important considerations to be taken when using it.

\subsection{Preliminaries}
Let $\imga, \imgb \in \R^{3 \times H \times W}$ be two images, where $H$ and $W$ denote the height and width.
We assume that there exists a non-empty subset of pixels in image $\imga$ that have corresponding pixels in image $\imgb$.
We refer to a single pixel in this subset as keypoint and denote it for $\imga$ as $\kpf{A}$ and the corresponding pixel on image $\imgb$ as $\kpf{B}$.

\textbf{View-Invariant Dense Descriptors.}
Let $f_\theta(\cdot)$ be a dense descriptor model that maps each pixel in an image $\mat{I}$ onto a $D$-dimensional latent space yielding a dense descriptor image $\mat{D} \in \R^{D \times H \times W}$.
Let $\dimga, \dimgb$ denote the descriptor images of $\imga, \imgb$, and $\desc{\kp{A}} = \dimga[x_A, y_A]\in\R^D$ the associated descriptor of $\kp{A}$, and respectively $\desc{\kp{B}}$ for $\kp{B}$.
The goal is to learn parameters $\theta$ such that $f_\theta(\cdot)$ will assign non-trivial, unique descriptors to two corresponding pixels, such that $\desc{\kp{A}}\approx\desc{\kp{B}}$, implying view-invariance, for example, with respect to scale, rotation, background, etc.

\textbf{Probabilistic Keypoint Heatmaps.}  
We can easily predict the location of a keypoint, given its descriptor, in a new image by finding the closest descriptor in latent space in $\dimgb$ with respect to $\desc{\kp{A}}$.
While this is sufficient for inference, one obtains the $(x,y)$-coordinates in a non-differentiable fashion, making it inadequate for training.
Instead, we compute a distance heatmap $\mat{H}^{\kp{A}\rightarrow B}$ over $\dimgb$ by taking the pairwise distances between $\desc{\kp{A}}$ and every descriptor of $\dimgb$, such that
\begin{align}
	\mat{H}_{xy}^{\kp{A}\rightarrow B} = \Delta(\desc{\kp{A}}, \dimg{B_{xy}}),
\end{align}
where $\Delta$ is some distance function, e.g., $\ell_2$-norm, or derived from a similarity measure, such as cosine similarity.
We assume cosine similarity and normalized descriptors in the following.
We obtain a probability distribution $P(x, y \mid \desc{\kp{A}} , \dimgb)$ by applying a temperature-scaled softmax function, such that
\begin{align}
	P(x,y \mid \desc{\kp{A}} , \dimgb) =
	\frac{
		\exp( H_{xy}^{\kp{A}\rightarrow B} / \tau )
	}{
		\sum_{i=1}^{H} \sum_{j=1}^{W} \exp( H_{ij}^{\kp{A}\rightarrow B} / \tau )
	},
\end{align}
where $\tau$ is the temperature.
By interpreting the expected values of the marginal distributions as coordinates, we derive $\kp{B}^\star = (x^\star, y^\star)$ as
\begin{align}
	x^\star & = \mu_x  = \textstyle \sum_{i=1}^{H} i \cdot \sum_{j=1}^{W} P(i, j \mid \desc{\kp{A}} , \dimgb), \label{eq:spat_exp_x} \\
	y^\star & = \mu_y = \textstyle \sum_{j=1}^{W} j \cdot \sum_{i=1}^{H} P(i, j \mid \desc{\kp{A}} , \dimgb). \label{eq:spat_exp_y} 
\end{align}
The variances $\sigma_{x}^2, \sigma_{y}^2$ follow naturally.
Conveniently, this formulation is differentiable.
If ground-truth annotations $\kpi{B}{(i)}$ exist, for example, in the case of pixelwise correspondences from 3D geometry, it is straight-forward to directly optimize the prediction error via the spatial expectation above, for example, with the loss function
\begin{align}
	\mathcal{L}_{\mathrm{distributional}, A\rightarrow B} & = \textstyle \sum_{i}^{N} \norm{\kpi{B}{\star~ (i)} - \kpi{B}{(i)}}_2, \label{eq:distr_loss}
\end{align}
where $N$ is the number of sampled keypoints in $\imga$.
The loss $\mathcal{L}_{\mathrm{distributional}}$ was previously introduced in a more general form in \cite{florence2020dense}.
A version relying on KL-divergence has also been proposed, see e.g., \cite{Vecerik2020}.

\subsection{Cycle-Correspondence Loss}
We now extend the above concept into a fully self-supervised training regime, when no ground-truth annotation $\kpi{B}{(i)}$ is given and we can not directly define an error to optimize as in Eq.~(\ref{eq:distr_loss}).
For sake of explanation, we temporarily assume the constraint that any $\kpi{A}{(i)}$ sampled has exactly one corresponding pixel $\kpi{B}{(i)}$ in $\imgb$, albeit unknown.
Given this assumption, we know that if the prediction $\kpi{B}{\star ~(i)}$ for $\imga\rightarrow\imgb$ is correct, then the associated descriptor $\desc{\kpi{B}{\star ~(i)}}$ should yield a prediction $\kpi{A}{\star ~(i)}$ from $\imgb\rightarrow\imga$, such that $\kpi{A}{(i)} \equiv \kpi{A}{\star ~(i)}$ holds.
This effectively completes a cycle of correspondence matching.
Since $\kpi{A}{(i)}$ is known, we can directly measure the prediction error, allowing us to define an error term for keypoint $i$ as
\begin{align}
	l_i & =  \norm{\kpi{A}{\star ~(i)} - \kpi{A}{(i)}}_2. \label{eq:vanilla_cycle_loss}
\end{align}
See Fig.~\ref{fig:cycle_overview} for a visualization.

\subsection{Implementation}
Although the loss is conceptually easy to formulate we now outline practical considerations that need to be taken into account for a successful implementation.

\subsubsection{Prevention of Short-Cut Learning}
To ensure the network will not ignore $\imgb$ and short-cut learn an identity mapping, we generate a copy $\imgaaug$ of input image $\imga$ and augment each separately.
As common in self-supervised training \cite{Chen2020,Grill2020,Florence2018,Adrian2022,graf2023learning,von2023treachery}, we apply a variety of augmentations to our input images.
In particular, we follow the selection presented in \cite{Adrian2022} by using affine transformations (rotation, scale), perspective distortion, and color jitter - the latter primarily for brightness and contrast augmentations.
We also know $\kpi{\hat{A}}{(i)}$, that is the location of $\kpi{A}{(i)}$ in $\imgaaug$, as the applied mapping is known, allowing us to redefine $l_i$ in Eq. (\ref{eq:vanilla_cycle_loss}) as
\begin{align}
l_i & =  \norm{\kpi{\hat{A}}{\star ~(i)} - \kpi{\hat{A}}{(i)}}_2. \label{eq:after_img_aug}
\end{align}

\subsubsection{Expected Descriptor and Keypoint Prediction}
In order to obtain $\desc{\kpi{B}{\star ~(i)}}$ in a differentiable fashion, we extend the concept of the spatial expectation, see Eq. (\ref{eq:spat_exp_x}, \ref{eq:spat_exp_y}), to compute the \textit{expected descriptor}, that is
\begin{align}
	\bar{\vc{d}}_{\kp{B}} & = \textstyle \sum_{i=1}^{H} \sum_{j=1}^{W}  \dimg{B_{ij}} \cdot P(i, j \mid \desc{\kp{A}} , \dimgb).
\end{align}
If the descriptors are normalized, one should additionally normalize $\bar{\vc{d}}_{\kp{B}}$, which we implicitly assume to be the case.
This allows us to define $P(x, y \mid \bar{\vc{d}}_{\kp{B}} , \dimg{\hat{A}})$ via $\bar{\vc{d}}_{\kp{B}}$ and determine $\kpi{\hat{A}}{\star ~(i)}$ using the spatial expectation.

\subsubsection{Handling Keypoints Without Correspondences}
\label{sec:kps_without_corr}
By training on unordered RGB images, objects may or may not be present, backgrounds change, or occlusion occurs.
Hence, we must now relax the above assumption that every $\kpi{A}{(i)}$ has a correspondence in $\imgb$.
Clearly, $l_i$ for some $\kpi{A}{(i)}$ without correspondence violates the underlying assumption of the cycle-consistency and calculated gradients might be completely counter-productive.
At the same time, as $\bar{\vc{d}}_{\kp{B}}$ is essentially a weighted sum of those descriptors in $\imgb$ most similar to $\desc{\kp{A}}$, the model could in practice still find a path from $\kpi{A}{(i)}$ to $\kpi{\hat{A}}{\star~(i)}$, even without a correspondence.
This short-cut learning should be prevented.

\begin{figure*}
	\centering
	\begin{subfigure}[b]{0.24\textwidth}
		\centering
		\includegraphics[width=\textwidth]{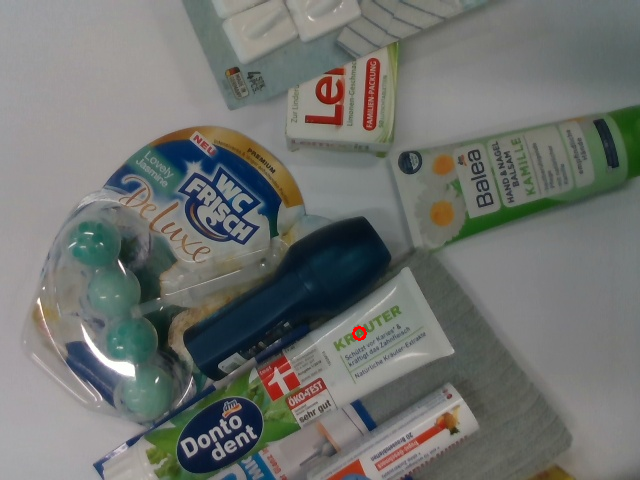}
	\end{subfigure}
	\begin{subfigure}[b]{0.24\textwidth}
		\centering
		\includegraphics[width=\textwidth]{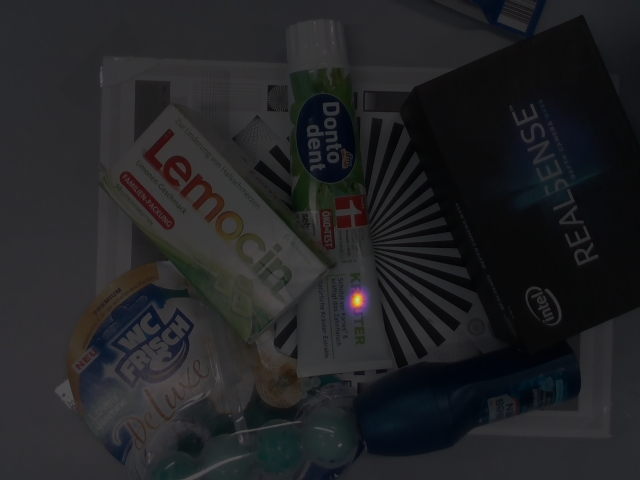}
	\end{subfigure}
	\begin{subfigure}[b]{0.24\textwidth}
		\centering
		\includegraphics[width=\textwidth]{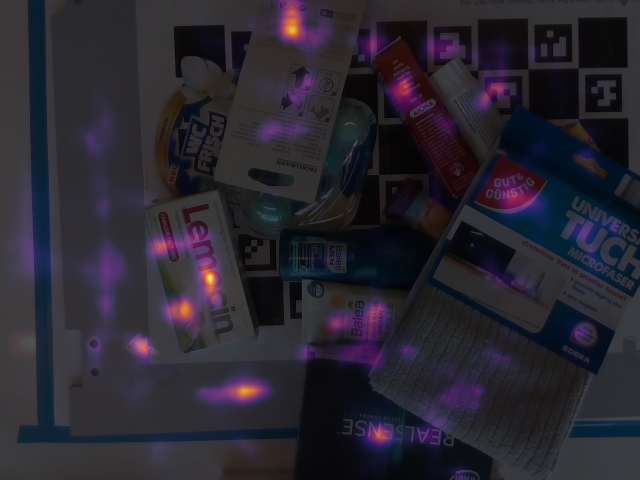}
	\end{subfigure}
	\begin{subfigure}[b]{0.24\textwidth}
		\centering
		\includegraphics[width=\textwidth]{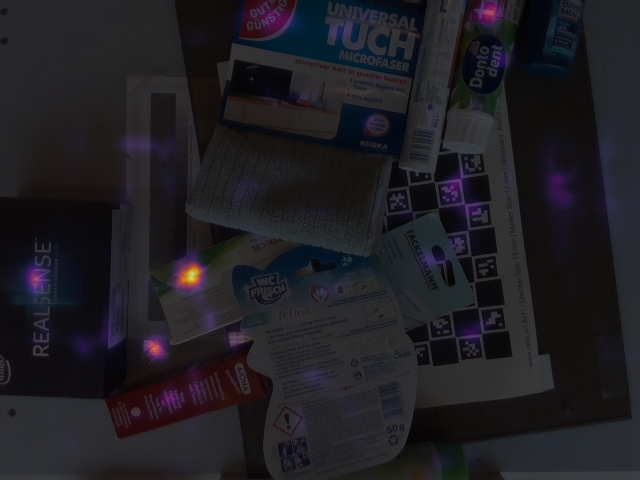}
	\end{subfigure}
	\caption{Visualization of the matching uncertainty.
		The red circle in the left most image marks the sampled keypoint.
		The following test images are superimposed with the predicted distribution as heatmap.
		If a correspondence exists (second from left), the mass of the distribution is well localized.
		If no correspondence exists (middle right and right most image), the mass is spread over various areas that are the most similar in descriptor space.
		Viewed best in color.
	}
	\vspace{-12pt}
	\label{fig:corr_non_corr_uncertainty}
\end{figure*}

We mitigate these issues by exploiting the previously determined probability distributions through two distinct mechanisms.
For both we first compute the summed variances $X_i = \chi_{\hat{A},i} + \chi_{B,i}$, with $\chi_{\cdot,i} = \sigma_{x,i}^2 + \sigma_{y,i}^2$, for the $i$-th keypoint predictions over images $\imgb$ and $\imgaaug$. 
Intuitively, we assume that $\chi_i$ is small, if a unique correspondence exists and the model is confident.
If no correspondence exists, or the model is not confident, $\chi_i$ should be large.
See Figure~\ref{fig:corr_non_corr_uncertainty} for a visualization of this emergent behaviour in our \CCL{} trained model.
For the first method we determine the $q$-quantile, e.g., $q=35\%$, over the $N$ summed variances $\{X_i\}_{i=0}^N$.
This gives us the $q\%$ most reliably detected points and we discard all other points from optimization.
For the second method, we modify Eq.~(\ref{eq:after_img_aug}), by scaling the contribution of each error with respect to the associated uncertainty, giving us the final loss
\begin{align}
	\mathcal{L}_\mathrm{cycle} & = \textstyle \sum_{i}^{N} \textstyle \frac{1}{1+X_i} \norm{\kpi{\hat{A}}{\star ~(i)} - \kpi{\hat{A}}{(i)}}_2, \label{eq:variance_weighted_loss}
\end{align}
where we add $1$ in the denominator to prevent the term from growing prohibitively large if some $X_i$ is smaller than $1$.
Importantly, we detach the calculated variances from the computational graph and do not back-propagate gradients, else the model will simply learn to make predictions with low confidence instead of solving the prediction task.

 \subsubsection{Pretraining \& Model Initialization}
 Although the model can be successfully trained from scratch, one can efficiently initialize by first performing a self-supervised pre-training akin to \cite{graf2023learning}.
 Here we directly match keypoints between $\imga$ and $\img{\hat{A}}$ by defining $P(x,y \mid \desc{\kp{A}} , \dimg{\hat{A}})$ and re-utilizing the distributional loss, such that
\begin{align}
	\mathcal{L}_\mathrm{identical} &= \mathcal{L}_{\mathrm{distributional}, A\rightarrow \hat{A}}, \label{eq:ident_loss}
\end{align}
 where descriptors are learned from correspondences generated synthetically via two augmented views, and each sampled $\kpi{A}{(i)}$ is guaranteed to be valid.
 A combined loss $\mathcal{L}=\mathcal{L}_{\mathrm{cycle}} + \lambda \mathcal{L}_{\mathrm{identical}}$ is also explored in the experiments.

\subsection{Relation to WarpC}
\label{sec:relation_warpc}
We note that \textit{WarpC} \cite{TruongWarpC} and its probabilistic extension \textit{PWarpC} \cite{TruongPWarpC} both utilize the notion of cycle-consistency in the context of dense matching.
Our work shares the same abstract concept of optimizing across unpaired images through completing some cycle, an idea also popularized in other contexts, e.g., in CycleGAN \cite{CycleGanJunYan}.
However, critical aspects differentiate our approaches.
Firstly, our optimization target is defined differently.
(P)WarpC implements the cycle concept by densely estimating the known ground-truth warp $W$ between $\imga$ and $\img{\hat{A}}$, induced by augmentations, by \textit{independently} predicting flow $F_{AB}$ between $\imga$ and $\imgb$ and $F_{B\hat{A}}$ between $\imgb$ and $\img{\hat{A}}$, such that $W \approx F_{AB} + F_{B\hat{A}}$.
In contrast, \CCL{} operates on a small subset of pixels.
For each we probabilistically estimate a descriptor over $\imgb$, which is directly used to infer a prediction over $\imgaaug$, making the prediction over $\imgaaug$ dependent on the prediction over $\imgb$.

Secondly, we differ from WarpC and PWarpC when it comes to discarding unmatchable pixels from optimization.
WarpC uses the current error between predicted flows and $W$ to compute a visibilty mask (cf. Eq. 9, \cite{TruongWarpC}).
PWarpC instead uses predicted confidence values to discard the $q\%$ most unrealiable points (cf. Eq. 9, \cite{TruongPWarpC}) like our first method.
In our work, we additionally scale the individual contribution of remaining error terms based on their associated confidence.
As we show in Sec.\ref{sec:ablation_quantile_drop}, this considerably improves our models performance, while making the exact choice of $q$ less sensitive.


\section{Experiments}
\label{sec:experiments}
We now discuss the methods and data of the evaluation setup.
This is followed by experimental results for the standard keypoint prediction accuracy task.
We then present a 6D grasp pose prediction experiment using a parallel gripper and conclude with an ablation study.

\subsection{Method Comparison}
We compare our loss primarily against task-agnostic methods for obtaining dense visual features.
We do, however, not review and compare against methodologies that focus on the data-generation side, such as sim-to-real DONs \cite{srdon} or NeRF-supervised DONs \cite{Yen-chen22}, nor methods that utilize already trained dense descriptor networks, such as \cite{von2023treachery}, as these can be combined with the presented \CCL{}.
Table \ref{tbl:main_results_dataset} summarizes all methods alongside their respective evaluation results.
The column \textit{(weakly) supervised} indicates which method requires, e.g., pixel-level masks or class labels.

The first set of methods we compare against are DON-like \cite{Florence2018} models.
(i) a model trained using augmented versions of a single image and extraction of synthetic correspondences (Identical View) according to \cite{graf2023learning} that utilizes only unordered RGB images like our method.
(ii) maskless multi-object scenes (MO-maskless) following \cite{Adrian2022}.
This is a specialized version of vanilla DONs utilizing ground-truth geometric correspondences extracted from RGBD sequences.
Finally, a fully supervised baseline: (iii) DON trained using synthetically composed collages of real image crops of objects (MO Collage Scenes) from many camera views, allowing for construction of object occlusions and other advanced compositions. 
This method uses both object-level masks and ground-truth geometric correspondences based on 3D scene reconstructions.
This yields a strong baseline setup to test impact of different levels of data complexity.
We trained all the variants using the distributional loss proposed in \cite{florence2020dense}.

We also compare against DINOv2 \cite{oquab2023dinov2}, a recent large-scale unsupervised trained vision approach.
We extract dense features using the authors provided code from last intermediate layer as it provided the best results out-of-the-box.

As closely related work, we also compare against WarpC \cite{TruongWarpC} and PWarpC \cite{TruongPWarpC}, both however specialized on dense matching via flow prediction.
These models are intended for dense geometric and semantic matching and seem to work best on images with large overlap or a single central object.

In addition to the vanilla version of \CCL{}, we also train a variant in combination with (Identical View), which shares the same data requirements.
Here we simply use $(\imga,\imgaaug)$ as input to $\mathcal{L}_{\mathrm{identical}}$ (Eq. \ref{eq:ident_loss}), while \CCL{} is trained as previously.
We combine both losses as $\mathcal{L}=\mathcal{L}_{\mathrm{cycle}} + \lambda \mathcal{L}_{\mathrm{identical}}$, where we found $\lambda=0.1$ to perform well.

\subsection{Datasets}
\label{sec:datasets}
We collected data of 12 objects in total to train and evaluate on, including challenging objects with transparent plastic, reflective, or black surfaces.
We provide method-specific training datasets described below, however, each method is compared against the same test dataset, which is described in more detail in Sec. \ref{sec:keypoint_prediction}.

\textbf{3D Reconstructed/RGBD-Datasets}: We followed the same protocol as in \cite{Florence2018} to collect RGBD-sequences using a wrist-mounted camera on a robot arm.
By means of 3D reconstruction, masks and geometric correspondences can be extracted.
This collection consists of 20 RGBD-sequences for training and five for validation.
Each sequence has around 480 frames.
This amount of sequences ensures that each object is seen from all sides and overall enough variety of scene and camera configurations that, e.g., also includes occlusions, is captured.
We trained \textit{MO-maskless}, \textit{MO Collage Scenes}, and WarpC on this dataset.
Although not relying on the annotations, WarpC would fail to train on the dataset discussed next.

\textbf{Unordered RGB}: For training \CCL{} an unordered collection of images is sufficient, for example, collected from a single, top-down view of a fixed camera.
We recorded 513 images, all from the same camera view, but altering the object arrangement in each frame.
To simplify this process, we obtained these by recording a continuous video stream, where an operator shuffles the objects and removes his hands from the camera view every other frame for a brief moment.
Duplicate static frames and blurry frames, e.g., where the operator hands are moving, can be trivially removed using common image processing tools.
We note that many of the frames, however, still contain the operator's hands, which we found did not hamper training success.
This way a complete training set was recorded in 5 minutes including processing.
This strongly contrasts the geometric datasets required for Dense Object Nets, which can take hours as they require multiple recordings per object, each taking several minutes, followed by 3D reconstructions and potentially manual mask generation.
We also trained PWarpC on this dataset as it yielded better results than on the above.

\begin{figure}
	\centering
	\includegraphics[width=\columnwidth]{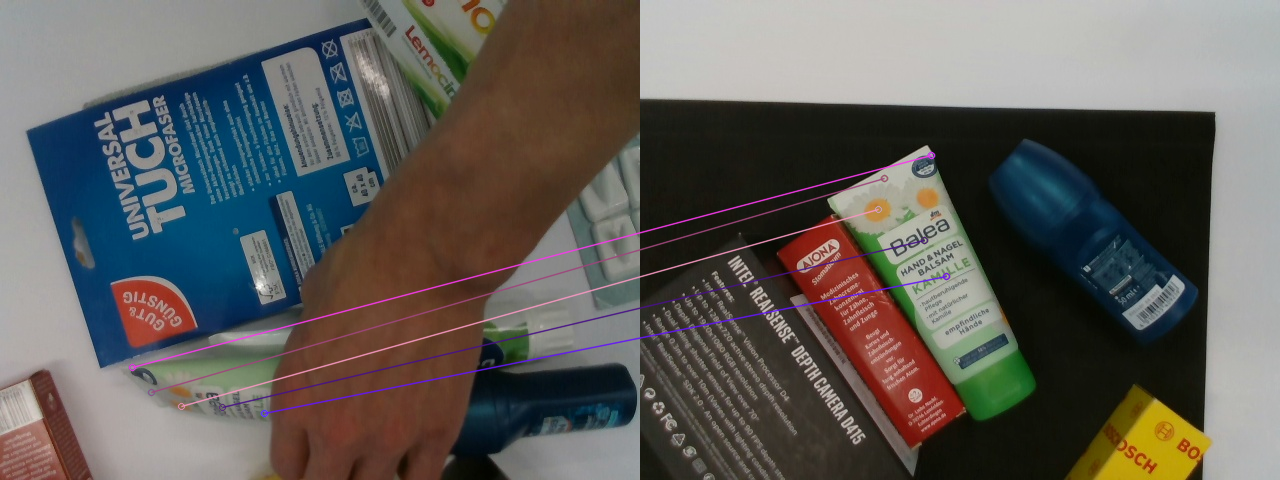}
	\caption{Example of hand-annotated, cross-scene keypoint matching test image pair.
		Occlusion, background changes, strong view-point or object pose changes are induced.
	}

	\label{fig:eval_test_example}
\end{figure}

\subsection{Training Details}

Similar to prior work \cite{Florence2018,Adrian2022,graf2023learning,HadjivelichkovK21}, we use a pretrained ResNet with an output stride of 8 and upsampling to match the input resolution, specifically a ResNet-50.
All input images are ImageNet normalized using $\mu=[0.485, 0.456, 0.406]$ and $\sigma=[0.229, 0.224, 0.225]$.
To increase efficiency we train using 16-bit precision using PyTorch \cite{Pytorch2019}.
For the \CCL{} trained model, we use the upsampled descriptor images only for evaluation, but for training the low resolution descriptor images are used.
This yields a descriptor image $\hat{\dimg{}} \in\R^{D \times \frac{H}{8} \times \frac{W}{8}}$, making the pairwise distance calculation very efficient.
We train with a batch size of 4 images and 2000 batches per epoch.
We sampled $N=500$ keypoint candidates per image pair or triplet.
The embedding size has been set to $D = 64$, following \cite{graf2023learning,von2023treachery}, and we use $\tau=0.03$, chosen by grid search.
Main results are reported for $q=35\%$.
We use AdamW \cite{LoshchilovH19} as optimizer with a fixed learning rate $lr=3\mathrm{e}{-5}$.
Models trained with \CCL ~have been initialized with the final checkpoint of the model obtained using identical view training \cite{graf2023learning}.

\subsection{Keypoint Prediction Accuracy}
\label{sec:keypoint_prediction}
Although the descriptors are task-agnostic, we follow a range of prior work \cite{Florence2018,Kupcsik2021,Yen-chen22,von2023treachery,srdon,Adrian2022} and evaluate how well keypoints can be matched across image pairs.
However, unlike some of aforementioned works we do not test using 3D reconstructed RGBD sequences for ground-truth annotation, as it limits testing the object poses and scene configurations of image pairs from static scenes.

Instead, we compiled a test dataset of 80 images, each depicting different scenes and object placements, and hand-annotated keypoints for each image and object.
In total 9124 image pairs, each featuring an object annotation consisting of around 10 keypoints on average.
Half the keypoints are located close to or on the object boundaries, the other half \textit{inside} the object.
This requires models to be robust to background changes and not calculate descriptors based on the background or close-by located objects.
Each image exhibits a different subset of objects, background changes, occlusion, or other scene composition factors, such as lighting conditions, see Fig. \ref{fig:eval_test_example} for an example.
This ensures that methods are robustly tested for their ability to generate descriptors that are view- and scene-invariant.

The results are compiled in Table \ref{tbl:main_results_dataset}.
We find that \textit{MO Collage Scenes} out-performs all methods, while relying on pixel-level masks and ground-truth geometric correspondences and thus having the highest data complexity.
\textit{\CCL{}} and \textit{MO-maskless} perform comparably, with the latter scoring higher on PCK@$\{3,5,10\}$ and \CCL{} on AUC and normalized mean pixel error.
The combination \textit{\CCL{}+Identical View} improves the results even further.

\textit{WarpC} and \textit{PWarpC} seem to struggle on our data.
When tested on image pairs from the same scene, that is same background and object arrangement but varied camera poses, they perform well.
However, when large parts of the images can not be matched and objects are subject to strong pose variations, as in our test set, the dense flow prediction is breaking down.
The pretrained \textit{DINOv2} model is not able to make accurate predictions under strong camera perspectives.
Although we found it can re-identify objects, it does not precisely locate positions.
This is partially also due to the large down-sampling factor of 14.

In summary, our proposed method outperforms all methods that do not rely on ground-truth geometric correspondences and approaches performance of the fully supervised MO Collage Scenes, despite being trained on only a small, but highly varied, unlabeled RGB-only dataset.

\begin{table*}[t!]
	\centering
	\scriptsize
	\tabcolsep=1.0mm
	\begin{center}
		\caption{
			Evaluation results for keypoint prediction.
			Methods requiring masks or, e.g., class labels (supervision) are marked.
			Metrics are percentage of correct keypoints (PCK@$k$), area-under-curve for PCK@$k$ for $k\in[1..50]$, and normalized mean pixel error.
			Standard deviation is denoted by the preceding $\pm$ symbol.
			Arrows $\uparrow$ and $\downarrow$ indicate if higher or lower is better.
		}
		\label{tbl:main_results_dataset}
		\begin{tabular}{lcccccccccc}
			\toprule
			 & (Weakly) & \multicolumn{5}{c}{PCK@}  &  AUC@  &   Norm.  Mean                                  \\
			Method  & Supervised & 3  $\uparrow$  & 5 $\uparrow$   & 10 $\uparrow$  & 25 $\uparrow$  & 50  $\uparrow$ & [1..50] $\uparrow$  &  Pixel Error $\downarrow$ \\
			\midrule

DINOv2 (b/14) pretrained \cite{oquab2023dinov2} & \--- & $.019\pm.135$ & $.05\pm.217$ & $.148\pm.356$ & $.368\pm.482$ & $.564\pm.496$ & $.151\pm.103$ & $.110\pm.137$\\
WarpC \cite{TruongWarpC} & \--- & $.04\pm.196$ & $.059\pm.236$ & $.082\pm.274$ & $.121\pm.326$ & $.173\pm.378$ & $.043\pm.109$ & $.247\pm.173$\\
Identical View  \cite{graf2023learning} & \--- & $.042\pm.202$ & $.100\pm.299$ & $.236\pm.425$ & $.442\pm.497$ & $.594\pm.491$ & $.177\pm.117$ & $.109\pm.145$\\
\midrule
\textbf{\CCL{} (ours)} & \--- & $.100\pm.300$ & $.222\pm.416$ & $.438\pm.496$ & $.664\pm.472$ & $.775\pm.418$ & $.266\pm.133$ & $.070\pm.129$\\
\textbf{\CCL{} (ours)} + Identical View \cite{graf2023learning} & \--- &
 \boldmath{$.124\pm.329$} &  \boldmath{$.261\pm.439$} &  \boldmath{$.481\pm.500$} &  \boldmath{$.690\pm.462$} &  \boldmath{$.793\pm.405$} &  \boldmath{$.277\pm.137$} &  \boldmath{$.064\pm.122$}\\

\midrule
\midrule

PWarpC \cite{TruongWarpC} & \cellcolor{red!35}\cmark & $.004\pm.067$ & $.013\pm.113$ & $.045\pm.208$ & $.165\pm.371$ & $.310\pm.463$ & $.066\pm.089$ & $.185\pm.165$ \\
MO-maskless \cite{Adrian2022} \cite{florence2020dense}  & \cellcolor{red!35}\cmark & $.130\pm.337$ & $.273\pm.445$ & $.476\pm.499$ & $.644\pm.479$ & $.741\pm.438$ & $.264\pm.133$ & $.071\pm.124$\\
MO Collage Scenes \cite{Florence2018,florence2020dense} & \cellcolor{red!35}\cmark & \boldmath{$.140\pm.347$} & \boldmath{$.289\pm.453$} & \boldmath{$.516\pm.500$} & \boldmath{$.700\pm.458$} & \boldmath{$.799\pm.401$} & \boldmath{$.286\pm.130$} & \boldmath{$.056\pm.110$}\\

			\bottomrule
		\end{tabular}
	\end{center}
	\vspace{-10pt}
\end{table*}

\subsection{Oriented Grasping Experiment}
\label{sec:robot_experiments}

\begin{table}
	\centering
	\caption{\textbf{Grasping Experiment Success Rate.}}
	\label{tbl:grasp_experiment}
	\begin{tabular}{l | c}
		\toprule
		Method / Loss                                            & Success Rate \\
		\midrule
		Identical View  \cite{graf2023learning}                 & 41.4\%       \\
		MO (maskless) \cite{Adrian2022}                                    & 44.8\%       \\
		MO Collage Scenes \cite{Florence2018,florence2020dense} & 77.6\%       \\
		\midrule
		\textbf{\CCL{} (ours)  }                                & 70.7\%       \\
		\bottomrule
	\end{tabular}
\end{table}

We compared the best performing methods on a 6D grasp pose prediction task using a parallel gripper as done in related work \cite{Adrian2022,Yen-chen22}.
To fairly compare the methods, we first recorded a single top-down view of each test object on a plain white background.
We define an axis along which we want to grasp by manually annotating two pixels per object and extracting the respective descriptors using each model.
We tested on six out of 12 training objects, as some would require a suction gripper.
We defined two alternative axis definitions per object, one with keypoints close to the object edges and one with locations further inside.
The latter is beneficial for methods trained without masks, like \cite{Adrian2022}, where descriptors are stable inside objects, but not close to the edges.
We test on cluttered scenes, where objects are placed on a heap with frequent background changes, including reflective surfaces and materials of similar color as the target object.
The current target object is always visible and graspable, but its placement might still induce strong perspective distortions w.r.t. the annotation image.
Each grasp configuration is tested on five different scene configurations, for a total of 30 grasps per network.
All networks are tested on the same scenes, which we accurately restore after each grasp attempt.

The results are compiled in Table~\ref{tbl:grasp_experiment}.
Unsurprisingly, the model trained on collage scenes has the most successful grasp attempts.
This model behaves less sensitive to changes in background due to strong background randomization and modeling of occlusion during training.
In comparison, the models (Identical View) and MO-maskless struggle, as they tend to integrate information from the background being trained on image pairs, where both images are from the same scene, as can be visualized by VisualBackProb \cite{visualbackprob}.
In contrast, \CCL, which is trained exclusively on RGB images showing different views, appears to learn more robustly encoded view and scene-invariant features, similar to the network trained on synthetically generated views.

\subsection{Ablation: Impact of Quantile Drop and Variance Scaling}

\label{sec:ablation_quantile_drop}
\begin{figure}
	\begin{center}
		\includegraphics[width=\columnwidth]{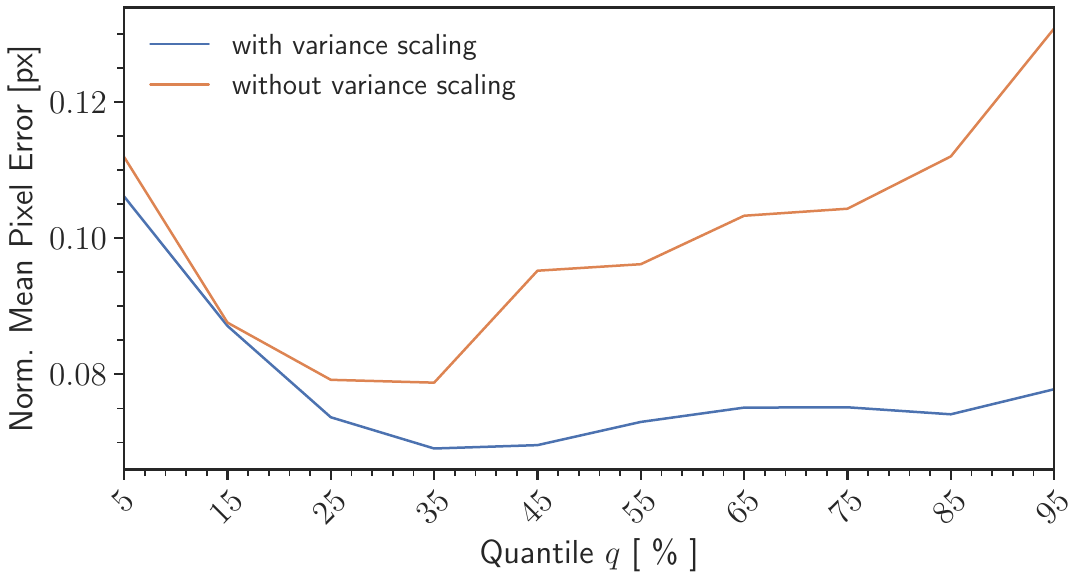}
	\end{center}
	\caption{Evaluation of prediction accuracy for different quantile $q$ and variance scaling.}
	\vspace{-10pt}
	\label{fig:abl_var_scale}
\end{figure}

We proposed two mechanisms to handle sampled keypoints candidates without correspondence in Sec. \ref{sec:kps_without_corr}.

To isolate their respective impact, different settings of $q$ were evaluated, with and without variance scaling.
See Figure \ref{fig:abl_var_scale} for results.
Clearly, having a smaller quantile $q$ helps prune bad samples efficiently.
However, particularly variance scaling leads to considerably better result overall while making the choice of the quantile $q$ much less sensitive.


\section{Limitations}
Despite the flexibility of the self-supervised formulation, some limitations need to be considered.
Firstly, the loss trains most effectively if both views have many valid pixel correspondences.
Although we demonstrated variance scaling and using a lower quantile threshold as remedy, we recommend to record data, e.g., as proposed.
Secondly, good performance of the loss will not necessarily imply good performance on downstream tasks, as our self-supervised loss is task agnostic.
Hence, validation directly on a downstream task or using a small labeled dataset for validation can prove helpful.

\section{Conclusions}
We presented a novel, self-supervised loss that allows to train complex dense visual feature extractors for object understanding in robotic manipulation using unordered collection of RGB images.
We effectively combine the benefits of pixel correspondence via alternative views and a simple data collection pipeline.
While there is still room for improvement, we could show highly competitive performance w.r.t. methods trained on registered RGBD scenes.
We plan to explore more advanced architectures, e.g., vision transformers, and methods for match cost calculation using self-attention in future work.

\section*{Acknowledgement}
We thank Christian Rauch, Christian Graf, and Miroslav Gabriel for their feedback and technical support.

\bibliographystyle{IEEEtran}
\bibliography{IEEEabrv,root}

\end{document}